\documentclass[conference]{IEEEtran}

\usepackage{cite}
\usepackage{amsmath,amssymb,amsfonts}
\usepackage{graphicx}
\usepackage{textcomp}
\usepackage{xcolor}
\usepackage{url}
\usepackage{booktabs}
\usepackage{enumitem}
\usepackage{orcidlink}
\usepackage{stfloats}

\begin{document}

\title{2.5-D Decomposition for LLM-Based Spatial Construction}

\author{\IEEEauthorblockN{Paul Whitten}
\IEEEauthorblockA{\textit{Rockwell Automation} \\
Mayfield Heights, OH, USA \\
\orcidlink{0000-0002-7787-473X}\,0000-0002-7787-473X}
\and
\IEEEauthorblockN{Li-Jen Chen}
\IEEEauthorblockA{\textit{Rockwell Automation} \\
Mayfield Heights, OH, USA \\
\orcidlink{0009-0000-8811-646X}\,0009-0000-8811-646X}

\and
\IEEEauthorblockN{Sharath Baddam}
\IEEEauthorblockA{\textit{Rockwell Automation} \\
Mayfield Heights, OH, USA \\
\orcidlink{0009-0008-8750-0548}\,0009-0008-8750-0548}
}

\maketitle

\begin{abstract}
Autonomous systems that build structures from natural-language instructions need
reliable spatial reasoning, yet large language models (LLMs) make systematic
coordinate errors when generating three-dimensional block placements. We present
a neuro-symbolic pipeline based on 2.5-D decomposition: the LLM plans in the
two-dimensional horizontal plane while a deterministic executor computes all
vertical placements from column occupancy, eliminating an entire class of errors.
On the Build What I Mean benchmark (160 rounds), GPT-4o-mini with this pipeline
achieves 94.6\% mean structural accuracy across 12 independent runs, within 3.0
percentage points of the 97.6\% ceiling imposed by architect-agent errors that
no builder-side improvement can address. This outperforms both GPT-4o at 90.3\%
and the best competing system at 76.3\%. A controlled ablation confirms that
2.5-D decomposition is the dominant contributor, accounting for 28.7 percentage
points of accuracy. The pipeline transfers directly to edge hardware: Nemotron-3
120B on an NVIDIA Jetson Thor AGX achieves 96.0\% mean structural accuracy with
the identical pipeline, slightly exceeding the cloud result.
Expanding the system prompt by four targeted examples to exceed the model's
8,320-token prefix cache page size, combined with low-effort reasoning, reduces
mean per-request latency by 3$\times$ to 19.7 seconds at 95.6\% accuracy. The
underlying principle, removing deterministic dimensions from the LLM's output
space, applies to any autonomous construction or assembly task where gravity or
other physical constraints fix one or more degrees of freedom. A transfer
experiment on 500 IGLU collaborative building tasks confirms the effect
generalizes beyond the primary benchmark.
\end{abstract}


\begin{figure*}[b]
    \centering
    \includegraphics[width=0.75\textwidth]{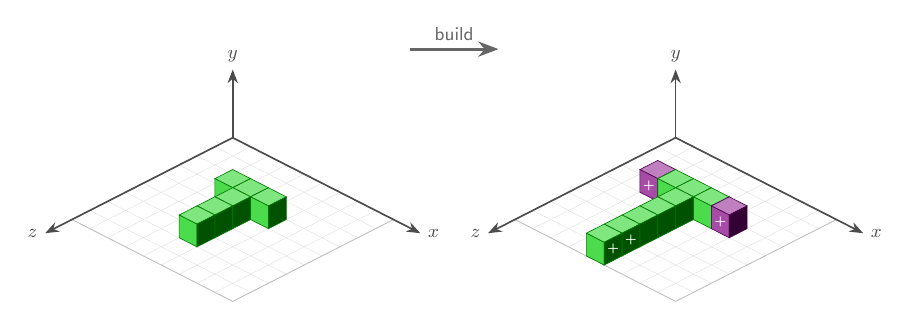}
    \caption{A benchmark round requiring T-shape recognition. Instruction:
    ``Keeping the T shape, extend the existing green structure by adding
    two green blocks to the longer base. Then add one purple block to each
    arm.'' New blocks are marked with~$+$.}
    \label{fig:tshape}
\end{figure*}

\section{Introduction}

Autonomous systems that construct physical structures from high-level
instructions must solve two problems: understanding what to build and
computing where to place each component. Large language models (LLMs)
are effective at the first problem but unreliable at the second. When
asked to produce three-dimensional coordinates, LLMs make systematic
errors in vertical placement, off-by-one stacking, and duplicate
positions \cite{yamada2024, bang2023}, consistent with LeCun's
observation that LLMs lack internal world models for enforcing physical
constraints \cite{lecun2022}.

We observe that many construction domains exhibit a 2.5-D structure: one
or more output dimensions are not free variables but deterministic
functions of the others and the current state. In gravity-constrained
block construction, the vertical coordinate of any new block is fully
determined by the column occupancy below it. The LLM does not need to
reason about this axis and in practice cannot do so reliably.

Our approach separates the problem accordingly. The LLM produces a plan
in the two-dimensional horizontal plane, specifying only $(x, z)$
positions, colors, and action types. A deterministic spatial executor
computes all vertical placement. This 2.5-D decomposition eliminates an
entire class of coordinate errors.
Fig.~\ref{fig:tshape} shows a representative round in which the agent
must recognize a T-shaped structure and extend it while preserving
symmetry.

The system includes four additional components: a structure analyzer that
detects geometric primitives in the existing grid, an underspecification
detector that decides when to ask a clarification question based on
expected-value analysis, an adaptive prompt enricher that injects
pattern-specific corrections before the LLM call, and a peephole plan
verifier that recognizes and rewrites erroneous output sequences after
generation.

An important ceiling on this benchmark is the architect agent that
answers clarification questions: it is itself an LLM and sometimes
provides incorrect information. Error analysis shows that 54.8\% of
residual failures originate in the architect, bounding the achievable
accuracy at approximately 97.6\% regardless of builder quality. Our
94.6\% result leaves only 3.0 percentage points of real headroom.

Our contributions are: (1) a 2.5-D decomposition that restricts the
LLM to horizontal planning while a deterministic executor handles
vertical placement, (2) an adaptive prompt enrichment methodology that
targets systematic LLM failure modes with pattern-matched corrections
injected before generation, paired with a peephole plan verifier that
recognizes and rewrites erroneous output sequences after generation,
(3) an empirical demonstration that this approach enables GPT-4o-mini
to outperform GPT-4o on block construction accuracy ($p < 0.0001$), and
(4) a decision-theoretic framework for underspecification handling under
an asymmetric scoring function.

\section{Related Work}

\textbf{LLM spatial reasoning.}
Yamada et al.\ \cite{yamada2024} benchmark LLMs on spatial tasks and
find that chain-of-thought prompting improves accuracy. Bang et al.\
\cite{bang2023} document persistent failures in relative positioning for
GPT-family models. Our work differs in that we do not attempt to improve
the LLM's spatial reasoning directly. Instead, we restrict the LLM to a
lower-dimensional output space and handle the eliminated dimension with
deterministic code.

\textbf{Plan-then-execute and neuro-symbolic decomposition.}
Wang et al.\ \cite{wang2023} show that separating planning from
execution improves zero-shot reasoning, and Khot et al.\ \cite{khot2023}
split complex tasks into sub-problems handled by specialized modules.
Yi et al.\ \cite{yi2018} decompose visual question answering into a
neural perception module and a symbolic execution engine. Our pipeline
adopts these ideas but extends them with a dimensional constraint: the
planning module operates in 2D while execution operates in 3D, a
separation not explored in prior decomposed-prompting work.

\textbf{LLM-guided robotic execution.}
Recent work constrains LLM outputs to what deterministic or learned
modules can execute. Ahn et al.\ \cite{saycan2022} ground language
model plans in robotic affordances, Liang et al.\ \cite{cap2022}
have the LLM generate policy code that calls perception APIs directly,
and Huang et al.\ \cite{innermonologue2022} close the loop with
environment feedback. Our approach shares this principle but exploits
a dimensional constraint specific to gravity-bound construction:
we remove an entire spatial axis from the LLM's output space and
compute it deterministically.

\textbf{3D construction agents.}
Wang et al.\ \cite{voyager2023} present VOYAGER, an LLM-powered
Minecraft agent for open-ended exploration. Zhu et al.\ \cite{gitm2023}
propose GITM for generally capable open-world agents. These systems
target sequential decision-making rather than precise structural
construction from underspecified instructions. The BWIM task requires
coordinate-level accuracy against a target configuration, making
coordinate arithmetic errors the dominant failure mode.

\section{Problem Formulation}

\subsection{Task Definition}

The BWIM benchmark \cite{bwim} defines a block construction task on a
discrete grid
$\mathcal{G} = \{0, \ldots, 8\} \times \{0, \ldots, 4\} \times
\{0, \ldots, 8\}$, where the axes correspond to width ($x$), height
($y$, vertical), and depth ($z$). Each cell is either empty or occupied by a
block of color $c \in \{\text{red}, \text{blue}, \text{green},
\text{orange}, \text{yellow}, \text{purple}\}$.

In each round, the builder agent receives a natural-language instruction
$I$ and a starting grid state $G_0$, and must produce a target grid
state $G^* = G_0 \cup B$ where $B$ is the set of new block placements.
The agent may issue one clarification question before building. The
scoring function awards $+10$ for a correct build, $-10$ for an
incorrect build, and $-5$ for each question asked
(see Fig.~\ref{fig:tshape} for an example round).

\subsection{2.5-D Decomposition}

The construction grid is a 2.5-D domain in the sense of
Marr \cite{marr2010}: the vertical axis is a deterministic function of
horizontal position and existing occupancy, analogous to 2.5-D machining
where the tool path is free in two axes but the third changes only in
discrete, computed steps \cite{held1991}. The grid enforces a gravity
constraint: a block at $(x, y, z)$ can only exist if $y = 0$ or a block
exists at $(x, y-1, z)$. The vertical coordinate of any new block is:
\begin{equation}
  y^*(x, z, G) = \min \{ y \in \{0, \ldots, 4\} \mid (x, y, z) \notin
  \text{dom}(G) \}
  \label{eq:ystar}
\end{equation}
This reduces the LLM's output space from
$|\mathcal{G}| \times |\mathcal{C}|$ to
$|\{0,\ldots,8\}|^2 \times |\mathcal{C}|$, eliminating $y$-coordinate
errors entirely.

\section{Architecture}

The agent processes each instruction through a six-stage pipeline:
\textbf{parse}, \textbf{analyze}, \textbf{plan}, \textbf{verify},
\textbf{execute}, and \textbf{format}. If any stage raises an
unrecoverable error, control falls back to a direct LLM call with an
engineered system prompt.

\begin{enumerate}
  \item \textbf{Instruction Parser.} Extracts the building directive
    from the incoming message and normalizes the representation.

  \item \textbf{Structure Analyzer.} Detects geometric primitives
    (rows, stacks, L-shapes, T-shapes) in $G_0$ and produces a
    structured description injected into the planner prompt.

  \item \textbf{Build Planner (LLM).} Decomposes the instruction into
    a plan $P = \langle a_1, \ldots, a_k \rangle$ of typed JSON actions,
    each specifying an action type, horizontal position $(x_i, z_i)$,
    color $c_i$, and count $n_i$. Nine worked examples in the system
    prompt \cite{wei2022} cover chains, L-shapes, T-shapes, and edge
    placements. No example includes $y$-coordinates.

  \item \textbf{Plan Verifier.} A rule-based module validates the plan
    against the instruction text and $G_0$ with four correction passes:
    direction consistency, endpoint cap correction, T-shape extend
    correction, and stacking plausibility.

  \item \textbf{Spatial Executor.} A deterministic engine processes each
    action on an in-memory grid, resolving relative references, computing
    $y$ via \eqref{eq:ystar}, and chaining positional context. No LLM
    call is involved. A same-color skip-forward rule in the extend
    handler detects occupied start positions and advances one grid step,
    preventing vertical stacking when horizontal extension is intended.

  \item \textbf{Response Formatter.} Translates the final grid state
    into the protocol-required output format.
\end{enumerate}

\begin{figure}[t]
    \centering
    \includegraphics[width=\columnwidth]{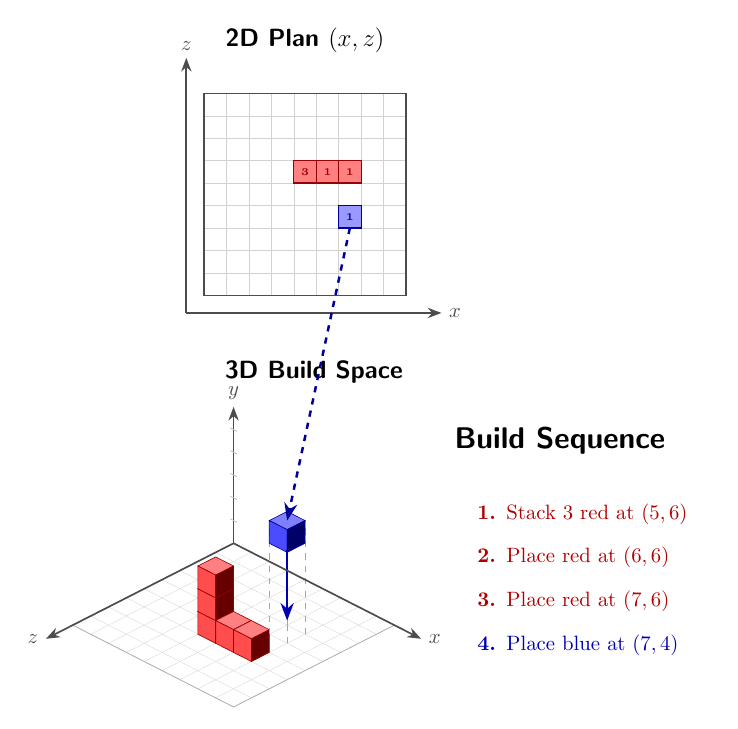}
    \caption{2.5-D decomposition: the LLM planner generates 2D plans
    with $(x, z)$ coordinates and action types. The deterministic
    executor computes vertical placement via column occupancy
    (Equation~\ref{eq:ystar}). Sequence order encodes vertical ($y$)
    position.}
    \label{fig:decomp}
\end{figure}

Fig.~\ref{fig:decomp} illustrates the decomposition. A plan specifying
``stack 3 red at $(5, 6)$, place red at $(6, 6)$'' contains no
$y$-coordinates. The executor computes a three-block column at $x=5$
(with $y = 0, 1, 2$) and a single block at $x=6$ (at $y=0$).

In our initial approach, the LLM produced the complete 3D output
directly, which required it to enumerate every block's $(x, y, z)$
coordinates and chain height calculations across stacking operations.
That approach suffered from frequent off-by-one $y$-errors, duplicate
placements, and miscounted heights. As shown in
Section~\ref{sec:results}, the current approach eliminates these failure
modes entirely.

\subsection{Underspecification Handling}

Many BWIM instructions deliberately omit color or block count. The
scoring function creates an asymmetric decision problem. Let $p$ denote
the probability of a correct guess:
\begin{align}
  \text{EV}_{\text{guess}}(p) &= 20p - 10 \label{eq:ev_guess} \\
  \text{EV}_{\text{ask}} &= 20p_a - 15 \label{eq:ev_ask}
\end{align}
Setting these equal with $p_a \approx 1$ yields the indifference
threshold $p^* = 3/4 = 0.75$. Below this threshold, asking dominates
guessing.

\textbf{Missing color.} Heuristic analysis determines whether a color
can be inferred from context (for example, reusing the sole color
mentioned) or is genuinely ambiguous. When ambiguous, the agent issues a
clarification question.

\textbf{Missing count.} Heuristic count inference (copying an adjacent
stack's height or defaulting to three) achieves approximately 65\%
accuracy on BWIM instructions, well below $p^*$, so the agent asks
whenever a count is underspecified and a question has not already been
used in the current round.

A complication is that the architect answering questions is itself an LLM
that receives only the target structure as coordinates, not the original
instruction. Generic questions such as ``How many blocks should be in the
stack?'' produce a 23\% error rate because the architect cannot identify
which stack is referenced. The agent generates color-specific questions
that name the block color, allowing the architect to identify the correct
stack by counting blocks of that color in its coordinate data.

Because only one question is permitted per round, the agent prioritizes
color over count (color errors are harder to recover from
heuristically). When multiple phrases lack counts, the remainder fall
back to a three-level heuristic cascade: copy the height of an adjacent
stack, use the tallest stack on the grid, or default to three (the modal
count in the benchmark, observed in 53\% of count-underspecified
trials).

\subsection{Adaptive Prompt Enrichment}

Certain spatial concepts (``each end,'' chain references, L-shape
extensions) are disproportionately error-prone for GPT-4o-mini. These
errors are systematic: specific input patterns reliably trigger specific
failure modes. We scan each instruction against 15 pattern-matching
rules before the LLM call. When a rule fires, a targeted correction
with a worked example is injected into the prompt. Rules are independent
and composable.

This methodology is domain-transferable: identify recurring failure
modes, classify by input trigger, write micro-corrections with worked
examples, inject dynamically via pattern matching, and validate against
regressions. It applies to any domain where LLMs make systematic,
input-predictable errors.

Table~\ref{tab:enrichment} lists five representative rules from the set
of 15. Each rule fires before the LLM call, matches a trigger phrase to
the systematic error the LLM produces, and injects a targeted correction
with a coordinate example into the prompt.

The plan verifier (stage 4) complements enrichment with post-generation
correction in the style of a compiler peephole optimizer
\cite{mckeeman1965}. Where a peephole pass matches known suboptimal
instruction sequences in generated code and replaces them, the plan
verifier matches known-bad patterns in the LLM's output plan and
rewrites them deterministically. Four correction passes handle direction
reversals, stale endpoint references, T-shape axis errors, and chain
reference coordinate drift. If critical errors remain after these
rewrites, the system triggers a full re-plan with diagnostic hints.

\begin{table*}[t]
\centering
\small
\caption{Representative adaptive prompt enrichment rules (5 of 15).}
\begin{tabular}{@{}lll@{}}
\toprule
\textbf{Trigger} & \textbf{LLM failure mode} & \textbf{Injected correction} \\
\midrule
``in front of'' & Keeps same $z$, stacks vertically on reference block & $z \mathbin{+}= 1$; $y$ resets to ground \\
``each end'' after extend & Uses pre-extension endpoints instead of post-extension & Recompute min/max after all modifications \\
L-shape extend & Extends through the junction, breaking the L & Extend away from junction at the far end \\
Chain ref (``the red one'') & References original block instead of just-built structure & Track position of most recent step \\
T-shape extend & Stacks vertically ($y\mathbin{+}=1$) instead of continuing stem & Identify stem axis, extend past base at $y\mathbin{=}0$ \\
\bottomrule
\end{tabular}
\label{tab:enrichment}
\end{table*}

\section{Experimental Setup}

We evaluate on the BWIM benchmark's 160-round scenario, which pairs the
builder agent with both a rational architect (which answers questions
correctly) and an unreliable architect (which sometimes provides
incorrect information). All runs use the same codebase and differ only in
model selection or component ablation.

\textbf{Models.} The primary model is GPT-4o-mini (2024-07-18). We
include GPT-4o (2024-08-06) as a comparison to assess whether model
scale provides additional gains beyond the deterministic pipeline. All
prompt engineering was developed against GPT-4o-mini. The GPT-4o runs
swap only the model identifier. All prompts, enrichment rules, temperature
settings, and pipeline logic are unchanged. We also evaluate
NVIDIA Nemotron-3-Super-120B-A12B \cite{nemotron2024}, an open-weight
model quantized to NVFP4, served locally via vLLM \cite{vllm2023} on an
NVIDIA Jetson Thor AGX Developer Kit \cite{jetsonthor} (14-core Arm
Neoverse V3AE CPU, 2560-core Blackwell GPU with fifth-generation Tensor
Cores). The baseline Nemotron runs use the identical 9-example system
prompt with no model-specific tuning. A follow-on latency experiment
expands the prompt to 13 examples to enable vLLM prefix caching and
enables the model's low-effort reasoning mode. Details are in
Section~\ref{sec:results}.

\textbf{Metrics.} We report structural accuracy, composite score
(accuracy-weighted with question penalties), and block-level F1.
Each round's predicted and target placements are compared as
sets of (color, $x$, $y$, $z$) tuples. F1 is the harmonic mean
of precision and recall over these sets, equivalent to the
Dice coefficient~\cite{dice1945}, providing graded partial credit
for near-miss rounds that binary accuracy marks as failures.
For GPT-4o-mini, we report statistics across 12 independent runs.
For GPT-4o and Nemotron-3, across six runs.

\textbf{Frozen codebase.} All reported runs use the same frozen code
commit. No code changes were made between runs within each model
condition. This ensures that variance reflects only LLM non-determinism,
not development changes. The pipeline calls the LLM at most twice per
round (planning plus at most one clarification re-plan).

\section{Results}
\label{sec:results}

\subsection{Comparison with Baselines}

\begin{table}[t]
\centering
\small
\caption{Comparison with baselines and competing systems.}
\begin{tabular}{@{}l@{\hskip 6pt}lcc@{}}
\toprule
\textbf{System} & \textbf{Model} & \textbf{Acc.} & \textbf{Score} \\
\midrule
BWIM baseline \cite{baseline} & None & 26.9\% & $-345$ \\
Ours without 2.5-D decomp. & 4o-mini & 65.9\% & $+29$ \\
hisandan/build-it-3 \cite{hisandan} & 4o & 76.3\% & $+765$ \\
\textbf{Ours (full)} & \textbf{4o-mini} & \textbf{94.6\%} & $\boldsymbol{+947}$ \\
\bottomrule
\end{tabular}
\label{tab:baselines}
\end{table}

Table~\ref{tab:baselines} shows our system compared with baselines and
the top competing system on the BWIM leaderboard.
The baseline agent returns a fixed build response regardless of the
instruction, establishing a floor.
The ``Ours without 2.5-D decomp.'' row is from the controlled ablation
(Table~\ref{tab:ablation}, $n = 7$): it retains the engineered fallback
prompt with spatial rules and underspecification questions but disables
the deterministic vertical-placement step, requiring the LLM to produce
full 3D coordinates directly with the few-shot examples expressed in 3D.
hisandan/build-it-3 \cite{hisandan} uses GPT-4o with example-anchored
prompting, speaker modeling, and strategic questioning for end-to-end
coordinate generation without a decomposition pipeline, improving from
49\% to 76.3\% across successive iterations ($n = 1$, score from the
public run log).
Our full system outperforms the competing system, which uses GPT-4o
(a larger and more expensive model), by 18.3 percentage points.

\subsection{Ablation Study}

\begin{table}[t]
\centering
\small
\caption{Ablation study ($n = 7$ per condition, GPT-4o-mini).}
\begin{tabular}{@{}lcccc@{}}
\toprule
\textbf{Configuration} & \textbf{Mean} & $\sigma$ & $\Delta$ & $p$ \\
\midrule
Full system ($n\!=\!12$) & 94.6\% & 0.72\% & -- & -- \\
$-$ 2.5-D decomp. & 65.9\% & 1.24\% & $-$28.7\,pp & $<$0.0001 \\
$-$ Underspec.\ qs & 76.7\% & 0.70\% & $-$17.9\,pp & $<$0.0001 \\
$-$ Enrichment & 83.3\% & 1.29\% & $-$11.3\,pp & $<$0.0001 \\
$-$ Skip-forward & 90.7\% & 1.22\% & $-$3.9\,pp & $<$0.0001 \\
$-$ Plan verifier & 94.0\% & 1.44\% & $-$0.6\,pp & 0.36 \\
\bottomrule
\end{tabular}
\label{tab:ablation}
\end{table}

Table~\ref{tab:ablation} presents an ablation of the final system.
Removing 2.5-D decomposition produces the largest accuracy drop
($-28.7$ pp, $p < 0.0001$), reducing accuracy to 65.9\%.
Underspecification questions contribute
$-17.9$ pp, adaptive prompt enrichment $-11.3$ pp, and the
skip-forward rule $-3.9$ pp, all significant at $p < 0.0001$. The plan
verifier (peephole correction) shows no significant effect ($-0.6$ pp,
$p = 0.36$), suggesting that the enrichment rules prevent most errors
that the verifier would otherwise need to correct.
Even with enrichment entirely disabled, the system reaches 83.3\%,
still 7 pp above the best competing system's 76.3\% with GPT-4o.

\subsection{Model Comparison}

\begin{table}[t]
\centering
\small
\caption{Model comparison on the frozen pipeline.}
\begin{tabular}{@{}lcccc c@{}}
\toprule
\textbf{Model} & $n$ & \textbf{Acc.} & $\sigma$ & \textbf{Score} & \textbf{F1} \\
\midrule
GPT-4o-mini & 12 & 94.6\% & 0.72\% & +947 & .987 \\
GPT-4o & 6 & 90.3\% & 0.52\% & +810 & .975 \\
Nemotron-3$^\dagger$ & 6 & 96.0\% & 0.31\% & +993 & .987 \\
\bottomrule
\multicolumn{6}{@{}l}{\footnotesize $^\dagger$Nemotron-3-Super-120B-A12B-NVFP4, Jetson Thor AGX.}
\end{tabular}
\label{tab:comparison}
\end{table}

Table~\ref{tab:comparison} compares GPT-4o-mini and GPT-4o on the
identical pipeline. GPT-4o-mini outperforms GPT-4o by 4.3 pp with
non-overlapping 95\% confidence intervals (Welch $t$-test:
$t = 14.3$, $\mathit{df} = 13.5$, $p < 0.0001$,
95\% CI $[3.6, 4.9]$ pp).
Block-level F1 exceeds 0.97 for all models, indicating that even
incorrect rounds are typically near-misses (a single misplaced block)
rather than catastrophic failures.
All prompt engineering was developed against GPT-4o-mini due to its
16$\times$ lower token cost. Each 160-round GPT-4o-mini run completes
in approximately 8.9 minutes at an estimated cost of \$0.07. GPT-4o
runs take approximately 10.1 minutes at \$1.10 per run, roughly
16$\times$ more expensive for lower accuracy on this pipeline. The cost
difference made GPT-4o impractical as the development model for iterative
prompt engineering across hundreds of test rounds. GPT-4o with the
untuned pipeline still reaches 90.3\%, well above the best competing
system's 76.3\% using GPT-4o without a decomposition pipeline,
indicating that the pipeline architecture provides value independent of
model-specific tuning.

\subsection{Edge Inference Latency}
\label{sec:latency}

The baseline Nemotron-3 evaluation used the identical 9-example pipeline
with full reasoning. Across six runs, mean accuracy was 96.0\%
($\sigma = 0.31\%$, +993 score), slightly exceeding GPT-4o-mini
(Welch $t = 5.9$, $p < 0.001$), with 59.0s mean latency (median 48.2s).
To reduce latency, a follow-on evaluation enabled low-effort reasoning
and expanded the system prompt from 9 to 13 examples (7,400 to 9,140
tokens). Nemotron-3 is a Mamba-2 hybrid. vLLM caches at the Mamba-2
page boundary (8,320 tokens), so exceeding it by 820 tokens achieved
85\% prefix cache hit rate and reduced mean TTFT from 4.0 to 1.1
seconds. Table~\ref{tab:latency} shows the result: 3.0$\times$ lower
latency at a 0.4 pp accuracy cost.

\begin{table}[t]
\centering
\small
\caption{Nemotron-3 per-request latency on Jetson Thor AGX.}
\begin{tabular}{@{}lcccc@{}}
\toprule
\textbf{Configuration} & \textbf{Median} & \textbf{Mean} & \textbf{P95} & \textbf{Acc.} \\
\midrule
Full reasoning & 48.2s & 59.0s & 122.9s & 96.0\% \\
Low-effort + caching & 17.1s & 19.7s & 37.0s & 95.6\% \\
\bottomrule
\end{tabular}
\label{tab:latency}
\end{table}

\subsection{Transfer to IGLU}

To test generality, we applied the 2.5-D decomposition to the IGLU
collaborative building dataset \cite{iglu2022}, an independent benchmark
with different instructions, a larger grid ($11 \times 9 \times 11$),
and no shared evaluation items with BWIM. We evaluated 500 gravity-compatible
tasks with GPT-4o-mini at temperature 0, comparing a bare coordinate-output
prompt against the 2.5-D decomposed prompt. The decomposition improved mean
block-level F1 from 0.723 to 0.798 (paired $t(499) = 5.76$, $p < 10^{-8}$,
$d = 0.26$, 95\% CI $[0.050, 0.101]$). No enrichment rules, plan verifier,
or underspecification handling were used. The result confirms that
dimensional reduction alone, the core architectural principle, transfers
to a structurally different spatial construction task.

\subsection{Error Analysis}

Across all 12 GPT-4o-mini runs (1,920 total rounds), 104 rounds produce
incorrect structures (5.4\% failure rate). Of these, 57 (54.8\%)
originate in the architect agent providing incorrect color information
in response to clarification questions. The remaining 47 (45.2\%) are
builder errors: 25 spatial reasoning mistakes and 22 incorrect
color-position assignments. Excluding architect errors, the builder
pipeline achieves 97.6\% accuracy (1,873/1,920 rounds correct).

\section{Limitations}

The primary evaluation uses a single 160-round benchmark (BWIM),
though Section~V-F confirms transfer to an independent dataset. The 2.5-D
decomposition depends on a gravity constraint that makes vertical
coordinates computable. Tasks without such a constraint would require a
different decomposition. The 15 adaptive enrichment rules were developed
by analyzing failure modes on the BWIM instruction set. While the
targeted failure patterns are general spatial reasoning problems, their
coverage of other instruction distributions is unknown. The competing
system's lower accuracy with GPT-4o (76.3\%) compared to our
GPT-4o-mini result (94.6\%) suggests genuine structural benefit beyond
benchmark-specific tuning. The error analysis shows that 54.8\% of
failures originate in the architect agent, creating a 3.0 pp accuracy
ceiling that builder-side improvements alone cannot close.

\section{Conclusion and Future Work}

We presented a neuro-symbolic approach to LLM-based block construction
in which the LLM plans in a two-dimensional horizontal plane while a
deterministic executor computes all vertical placement. On the BWIM
benchmark, this 2.5-D decomposition enables GPT-4o-mini to achieve
94.6\% mean structural accuracy, outperforming GPT-4o at 90.3\%
($p < 0.0001$). A controlled ablation study confirms that 2.5-D
decomposition accounts for the largest single contribution ($-28.7$ pp
when removed).

The broader principle is that LLM spatial reasoning can be made reliable
by identifying which dimensions of the output space are deterministically
computable from task constraints and removing those dimensions from the
LLM's responsibility. This principle applies to any domain where one or
more output dimensions are deterministic functions of the others, a
condition we term 2.5-D structure by analogy with Marr's 2.5-D
sketch \cite{marr2010}.

This decomposition strategy applies wherever physical laws constrain one
or more output dimensions, including gravity-bound robotic assembly,
terrain-following path planning, and automated equipment maintenance.
Identifying and removing deterministic dimensions from the LLM's output
space offers a practical path toward reliable autonomous operation in
physically constrained environments.

Future work includes developing methods for the builder to detect
architect errors and investigating whether the 2.5-D principle extends
to tasks with structural stability or connectivity constraints.

\section{Acknowledgement}

We are greateful to Rucha Apte and NVIDIA for their assistance in optimizations
suggestions for edge inference on the Jetson AGX Thor in helping to to realize
its potential.

\section*{Reproducibility}

Source code, prompts, enrichment rules, evaluation scripts, and scoring
logs are available at
\url{https://github.com/paulwhitten/AgentWhetters-bwim} (tag
\texttt{v1.0.5}). The IGLU transfer experiment code is in
\texttt{iglu/} on the \texttt{main} branch of the same repository. Models: \texttt{gpt-4o-mini-2024-07-18},
\texttt{gpt-4o-2024-08-06} (OpenAI API), and
\texttt{Nemotron-3-Super-120B-A12B-NVFP4} (local, vLLM~\cite{vllm2023},
Jetson Thor AGX Developer Kit, Blackwell GPU, Arm Neoverse V3AE CPU).
Build planner temperature~0.1. All other calls
temperature~0.2. The benchmark's architect (green) agent uses
\texttt{gpt-4o-mini}.

\bibliographystyle{IEEEtran}
\bibliography{references}

\end{document}